\documentclass{article}
\usepackage{ijcai25}

\usepackage{times}
\usepackage{soul}
\usepackage{url}
\usepackage[hidelinks]{hyperref}
\usepackage[utf8]{inputenc}
\usepackage[small]{caption}
\usepackage{graphicx}
\usepackage{amsmath}
\usepackage{amsthm}
\usepackage{booktabs}
\usepackage[switch]{lineno}

\usepackage{amssymb}
\usepackage{multirow}
\usepackage{booktabs}
\usepackage{makecell}
\usepackage{textcomp}
\usepackage[linesnumbered,ruled]{algorithm2e}

\SetCommentSty{mycommfont}
\usepackage{algpseudocode}
\SetKwInput{KwInput}{Input}                
\SetKwInput{KwOutput}{Output}              

\title{Hybrid Mesh-Gaussian Representation for Efficient Indoor Scene Reconstruction}

\author{
Binxiao Huang$^{1}$
\and
Zhihao Li$^{2}$
\and
Shiyong Liu$^{2}$
\and
Xiao Tang$^{2}$
\and
Jiajun Tang$^{3}$ 
\and \\
Jiaqi Lin$^{4}$ 
\and 
Yuxin Cheng$^{1}$
\and
Zhenyu Chen$^{2}$
\and
Xiaofei Wu$^{2}$
\and
Ngai Wong$^{1\dag}$
\affiliations
$^1$The University of Hong Kong \\
$^2$Huawei Technologies Ltd  \\
$^3$Peking University  \\
$^4$Tsinghua University\\
\emails
\{bxhuang, yxcheng, nwong\}@eee.hku.hk, \{zhihao.li, liushiyong3, tangxiao12, wuxiaofei2\}@huawei.com, jiajun.tang@pku.edu.cn, \\
linjq22@mails.tsinghua.edu.cn, zhenyu.chen@tum.de
}

\begin{document}

\maketitle
\renewcommand{\thefootnote}{$\dag$} 
\footnotetext{~Corresponding author.} 

\begin{abstract}
3D Gaussian splatting (3DGS) has demonstrated exceptional performance in image-based 3D reconstruction and real-time rendering. However, regions with complex textures require numerous Gaussians to capture significant color variations accurately, leading to inefficiencies in rendering speed. To address this challenge, we introduce a hybrid representation for indoor scenes that combines 3DGS with textured meshes. Our approach uses textured meshes to handle texture-rich flat areas, while retaining Gaussians to model intricate geometries. The proposed method begins by pruning and refining the extracted mesh to eliminate geometrically complex regions. We then employ a joint optimization for 3DGS and mesh, incorporating a warm-up strategy and transmittance-aware supervision to balance their contributions seamlessly.Extensive experiments demonstrate that the hybrid representation maintains comparable rendering quality and achieves superior frames per second FPS with fewer Gaussian primitives.


\end{abstract}    
\section{Introduction}
\label{sec:intro}


Reconstructing a high-quality 3D representation with multiple unordered images is a critical task in computer vision and graphics. In recent years, an implicit representation approach called the neural radiance field (NeRF)~\cite{DBLP:conf/eccv/MildenhallSTBRN20} shows extraordinary performance. NeRF combines the deep learning and volumetric rendering approach to produce high-fidelity images from novel views.
However, the requirement of dense sampling of spatial positions along rays heavily slows down the training and rendering speed. By representing the scene with plenty of anisotropic 3D Gaussian primitives, 3D Gaussian splatting (3DGS)~\cite{DBLP:conf/eccv/MildenhallSTBRN20} has emerged as a charming and efficient scene representation that achieves real-time rendering for scenes with unprecedented fidelity. 3DGS parameterizes the scene as an optimized set of 3D Gaussians obtained from the structure from motion or randomly initialized. It efficiently renders all relevant Gaussian primitives into a 2D plane via splatting-based rasterization to obtain 2D images. \cite{DBLP:journals/corr/abs-2402-01459,DBLP:journals/corr/abs-2402-04796,DBLP:journals/corr/abs-2406-07499} utilize the GS to yield accurate reconstructions to extract meshes to simulate physical interactions. 
Despite the compelling performance achieved by 3D-GS, the number of Gaussian primitives is excessively redundant~\cite{DBLP:journals/corr/abs-2406-07499,DBLP:conf/eccv/FangW24}. For tile-wise sorting and pixel-wise alpha-blending in the splatting-based rasterization, massive Gaussian primitives can seriously slow down rendering. Several methods have explored the contributions~\cite{DBLP:journals/corr/abs-2406-07499,DBLP:journals/corr/abs-2311-17245} or important scores~\cite{DBLP:journals/pacmcgit/PapantonakisKKLD24} of each Gaussian to prune the redundant Gaussian for a compact representation. These methods focus on Gaussian attributes to reduce Gaussians. This paper proposes leveraging a GS-mesh hybrid representation to reduce Gaussians. For rich-texture flat areas, vanilla GS requires a large number of Gaussians to offer a precise representation, while textured mesh is capable of expressing high-frequency information through texture. Specifically, we utilize textured mesh (with an opacity set to 1) as the background with certain depths to reduce the number of Gaussians and accelerate the inference rendering speed.

\begin{figure}[htp]
    \centering
    \includegraphics[width=0.98\linewidth]{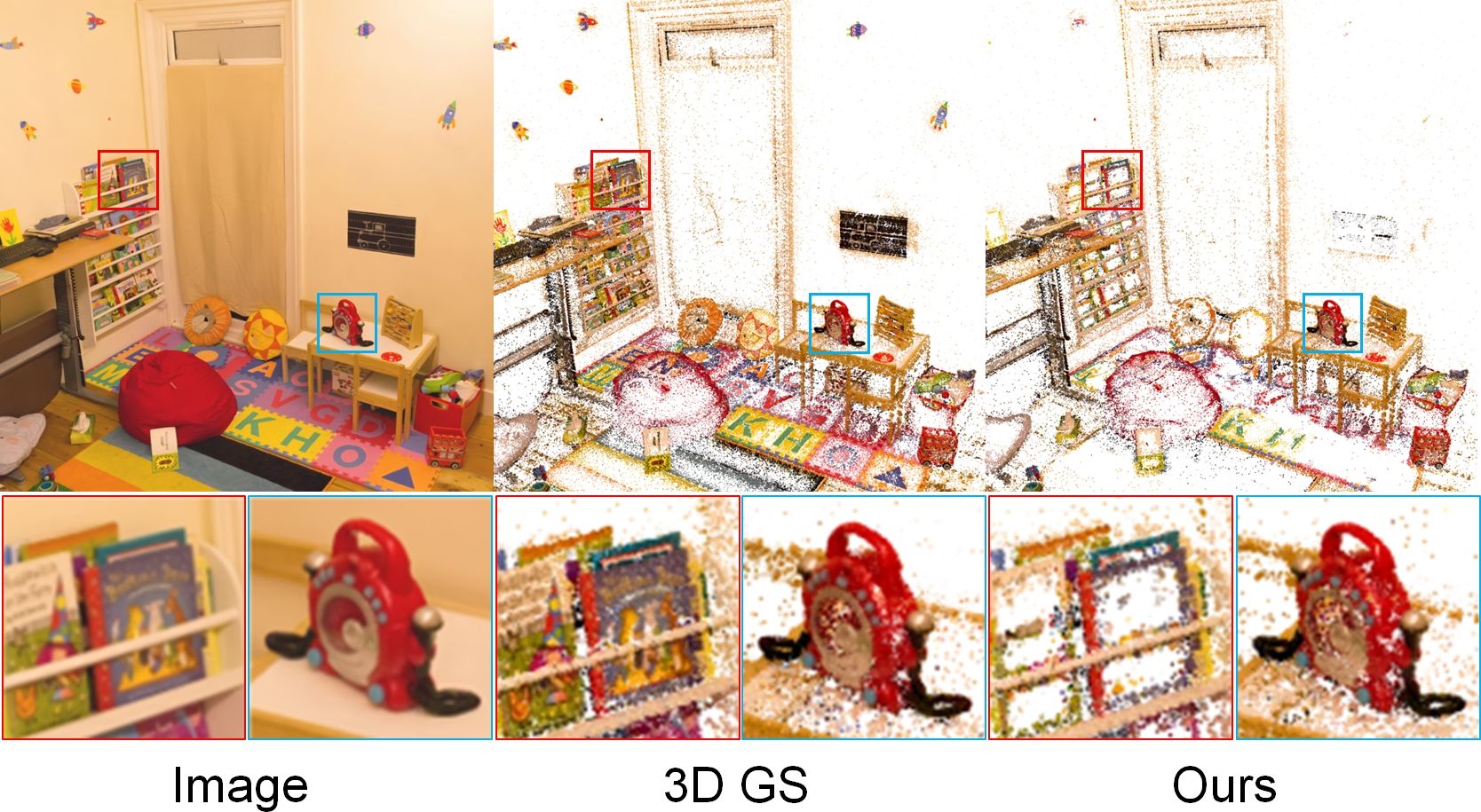}
    \caption{Visualization of Gaussian distributions and corresponding rendered images. The hybrid representation significantly reduces the number of Gaussians in texture-rich regions while maintaining Gaussian density in geometrically complex areas.}
    \label{fig:motivation}
\end{figure}

As the distribution of 3D Gaussians shown in Fig.~\ref{fig:motivation}, the Gaussians mainly concentrate on complicated geometry and rich texture regions. To precisely represent complex geometries, a substantial number of fine-grained and elaborate meshes are required~\cite{chen2023text2tex,chen2024scenetex}, which are not obtainable in real-world scene captures. Consequently, we still rely on the Gaussians to represent the complicated regions accurately. For texture-rich areas, we use meshes to represent flat geometric structures and texture maps to express high-frequency content information to reduce the number of Gaussian primitives significantly. Nevertheless, an ideal mesh is not available in real-world scenes. Various methods have been designed to extract mesh from multiple images using NeRF~\cite{DBLP:conf/eccv/MildenhallSTBRN20} or 3D-GS~\cite{DBLP:journals/tog/KerblKLD23}. Owing to the weak supervision over texture-less areas and the challenges in representing geometrically complex regions, the extracted mesh exhibits noticeable geometric errors. We design some metrics to clear inaccurate and fine-grained meshes, preventing them from misleading the 3D representation. Then, we integrate the textured mesh into the 3DGS method to jointly represent the 3D indoor scenes. To reduce the Gaussians in front of the mesh, we employ texture supervision to narrow the visual disparity between the images rendered from the mesh and the ground truth. Additionally, we introduce a transmittance-aware mask to prevent the projection of front objects' colors onto the mesh. Compared to 3DGS, the proposed hybrid approach significantly reduces Gaussian primitives with a comparable quality. 

In summary, our contributions are as follows:
\begin{itemize}
    \item We propose a hybrid representation that leverages the advantages of Gaussian splatting and textured mesh to reduce the Gaussian primitives.
    \item We devise certain metrics to prune the flawed meshes in geometrically complex areas and retain them in texture-rich flat regions to facilitate joint rendering.
    \item We introduce transmittance-aware supervision to facilitate joint optimization of the hybrid representation.
\end{itemize}

\section{Related Works}
\label{sec:related}

\subsection{3D Gaussian Splatting}
Recently, 3DGS~\cite{DBLP:journals/tog/KerblKLD23} has emerged as an advancement in novel view synthesis, achieving real-time, high-fidelity rendering. Different from implicit neural fields with volume rendering based on ray marching, 3DGS represents the 3D scene with explicit Gaussians that can be rendered through tile-based sorting and rasterization processes. A plethora of subsequent studies have surfaced, innovating technology across diverse fields such as human avatar rendering~\cite{DBLP:journals/corr/abs-2402-10483,DBLP:conf/cvpr/LiuZTSZLLL24}, content creation~\cite{DBLP:journals/corr/abs-2404-03575,DBLP:conf/eccv/ZhouFXCCBYWK24},
and scene rendering~\cite{DBLP:conf/cvpr/WuYFX0000W24,lin2025decoupling,DBLP:conf/cvpr/LinLTLLLLWXYY24}.

However, the performance of 3DGS models is constrained by the large number of Gaussians due to the rasterization process. ~\cite{DBLP:journals/corr/abs-2404-06109} leverages a per-pixel error function as the density criterion and introduces a mechanism to control the total number of Gaussians. Taming 3DGS~\cite{DBLP:journals/corr/abs-2406-15643} designs a steerable densification method that yields any desired number of Gaussians after training. ~\cite{DBLP:journals/pacmcgit/PapantonakisKKLD24} assigns a redundancy score for each Gaussian to measure how necessary a Gaussian is to represent the scene. Then, Gaussians whose redundancy scores are higher than the adaptive threshold are pruned. Mini-splatting~\cite{DBLP:conf/eccv/FangW24} proposes an intersection-preserving technique to retain Gaussians that act as intersection points and adopts an importance-weighted sampling approach to maintain a sparse set of Gaussians. LightGaussian~\cite{DBLP:journals/corr/abs-2311-17245} computes a global significance score of each Gaussian considering the Gaussian's volume, opacity, transmittance, and hit count of each training view. It then ranks all Gaussians by their global significance scores to quantitatively guide the pruning of the lower-ranked Gaussians. All these methods rely solely on GS to represent the 3D scene and take into account the attributions of Gaussians to prune the unimportant ones. As illustrated in Fig.~\ref{fig:motivation}, the Gaussian primitives predominantly concentrate on the texture-rich and geometrically complex regions. Our hybrid representation aims to reduce the Gaussians representing the texture-rich area via textured mesh while behaving the same as 3DGS in regions with intricate geometry. Theoretically, our method is perpendicular to previous approaches.



\subsection{Hybrid representation}

In order to combine the benefits of multiple 3D expressions (\textit{i.e.} NeRF, mesh, GS), various hybrid approaches have been proposed~\cite{DBLP:conf/cvpr/Wen0RSW24,DBLP:conf/eccv/DhamoNMSSZP24,DBLP:journals/corr/abs-2408-10041}.
VMesh~\cite{DBLP:conf/siggrapha/GuoC0HSZ23} depicts an object with a textured mesh and an auxiliary sparse volume for efficient view synthesis. HybridNeRF~\cite{DBLP:conf/cvpr/TurkiABPKRZR24} combines the best of surface and volume-based rendering into a single model to achieve a real-time rendering with high quality. SuGaR~\cite{DBLP:conf/cvpr/GuedonL24} regulates the 3D Gaussians to align well with the surface to extract the mesh and jointly optimizes the Gaussians and the mesh. GSDF~\cite{DBLP:journals/corr/abs-2403-16964} and NeuSG~\cite{DBLP:journals/corr/abs-2312-00846} improve the quality of surface reconstruction by combining the benefits of 3DGS with neural implicit fields.
MeshGS~\cite{choi2024meshgsadaptivemeshalignedgaussian} introduces several regularization techniques to precisely bind
Gaussian splats with the extracted mesh surface for high-quality rendering. Some methods use a GS-mesh hybrid representation to make the GS editable and controllable. GaMeS~\cite{DBLP:journals/corr/abs-2402-01459} parameterizes each Gaussian primitive by the vertices of the mesh face to allow modifying the Gaussians in a mesh manner. SplattingAvatar~\cite{DBLP:conf/cvpr/ShaoWLWL00024} disentangles the motion and appearance of an avatar with explicit mesh faces and implicit appearance modeling with GS. It can directly control the mesh to empower its compatibility with animation techniques. GaussianAvatars~\cite{DBLP:conf/cvpr/QianKS0GN24} places Gaussian splats onto a parametric morphable face mode to enable control in terms of expression, pose, and viewpoint of head avatars. HAHA~\cite{DBLP:journals/corr/abs-2404-01053} models human avatars utilizing Gaussian splatting and a textured mesh for efficient rendering with much fewer Gaussians. This method assumes the availability of a water-tight SMPL-X parametric mesh and doesn't need to consider the occlusion of objects in the scene. Kim~\emph{et.al}
~\cite{DBLP:journals/corr/abs-2407-16173} integrates mesh to represent the room layout and employs GS for other objects with a prior segmentation. Nevertheless, the textured mesh is only used for layouts that demand minimal Gaussians, and the manual acquisition of the mesh from a synthetic dataset makes it impractical for real-world scenes. Our method leverages mesh to dominate the texture-rich area, effectively reducing the number of Gaussian primitives while maintaining a comparable rendering performance.
\section{Methodology}
\label{sec:method}

\begin{figure*}
    \centering
    \includegraphics[width=0.95\linewidth]{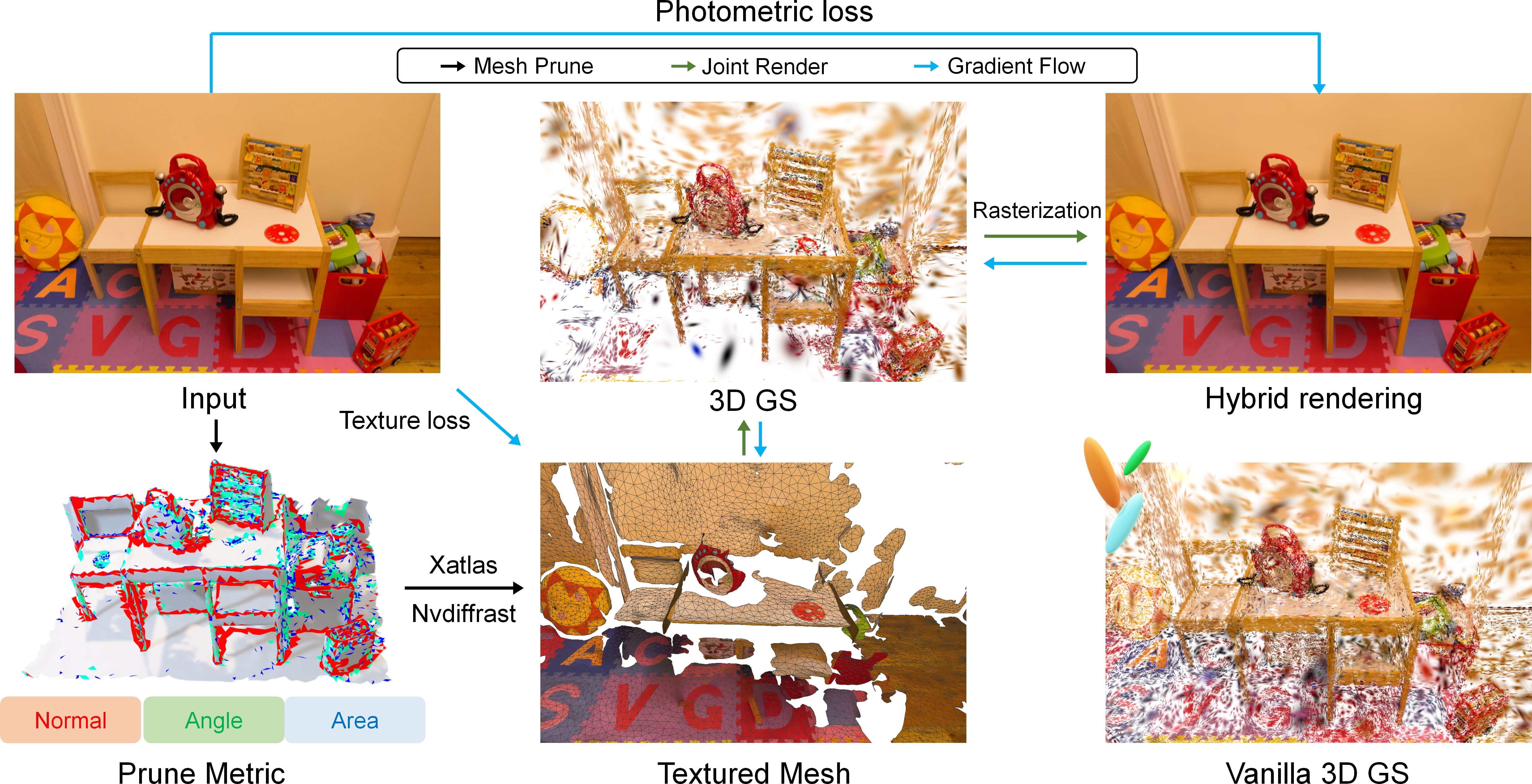}
    \caption{Overview of the proposed method pipeline. We first use normal, angle, and area size metrics to remove the meshes representing the geometrically complex region. Then, we use hybrid representation to induce Gaussians to fill the empty space and inhibit its densification in texture-rich flat regions under photometric and texture supervision.}
    \label{fig:pipeline}
\end{figure*}

Taking multi-view images with calibrated camera poses and sparse point clouds derived from the structure from motion (SFM) algorithm as input, we aim to combine the 3DGS and textured mesh to accurately represent indoor scenes while maintaining rendering quality and improve efficiency.

\subsection{Preliminary}

\paragraph{3D Gaussian Splatting} 3DGS~\cite{DBLP:journals/tog/KerblKLD23} utilizes anisotropic 3D Gaussian primitives to represent the 3D scene, achieving state-of-the-art visual quality and rendering efficiency. Each explicit Gaussian primitive is characterized by the following contributions: position center $\boldsymbol{u} \in \mathbb{R}^3$, opacity $\alpha$, orthogonal rotation matrix $\boldsymbol{R}$, diagonal scale matrix $\boldsymbol{S}$, and spherical harmonics (SH)  coefficients. Each Gaussian is defined by $\boldsymbol{u}$ and covariance matrix $\Sigma \in \mathbb{R}^{3 \times 3}$ as:
\begin{equation}
    G(\boldsymbol{x}) = exp(-\frac{1}{2}(\boldsymbol{x}-\boldsymbol{u})^T \Sigma ^{-1}(\boldsymbol{x}-\boldsymbol{u})).
\end{equation}

To guarantee the physical meaning of the covariance matrix during optimization, it is formulated as $\Sigma = \boldsymbol{RSS^TR^T}$ to keep a positive semi-definite character. To render images, 3D Gaussian primitives are projected onto the 2D image plane with the Jacobian affine approximation $\boldsymbol{J}$ of the project matrix and view transformation $\boldsymbol{W}$: $\Sigma' = \boldsymbol{JW}\Sigma\boldsymbol{W^TJ^T}$. Then alpha blending is applied from front to back based on the sorted depth to render the color of each pixel as follows:
\begin{equation}
    C(\boldsymbol{p}) = \sum_{i\in\mathcal{N}} c_{i}\sigma_{i} \prod^{i-1}_{j=1}(1-\sigma_j), \quad \sigma_i = \alpha_i G_{i}'(\boldsymbol{p}),
\end{equation}
where N denotes the number of sorted Gaussians related to the pixel $\boldsymbol{p}$, $c_i$ denotes the color of the projected 2D Gaussian $G_{i}'$. The $\mathcal{L}_1$ loss and D-SSIM term between the ground truth image $\boldsymbol{I}$ and the rendered image $\hat{\boldsymbol{I}}$ is utilized to optimize 3D Gaussians primitives:
\begin{equation}
\label{gs_loss}
    \mathcal{L}_{c}(\boldsymbol{I}, \hat{\boldsymbol{I}})= (1 - \lambda) \mathcal{L}_1(\boldsymbol{I}, \hat{\boldsymbol{I}}) + \lambda \mathcal{L}_{D-SSIM}(\boldsymbol{I}, \hat{\boldsymbol{I}}),
\end{equation}
where $\lambda$ is set to 0.2 as default.

\paragraph{Textured mesh} A mesh is a collection of vertices and faces that provides the geometric description of 3D models. A texture map is an image that maps the detailed visual information of images onto the surface of a mesh. To render the triangle meshes with a texture map, barycentric interpolation is applied to accurately assign texture colors to individual pixels in the rendered image based on UV coordinates. 
In this way, textured meshes enable fast real-time rendering of flat regions with realistic visual results. However, it struggles to model intricate structures, which we rely on the Gaussians to represent.

\subsection{Overview}
Our hybrid representation depicts indoor scenes with triangular mesh surfaces and anisotropic Gaussian primitives. In this section, we first clear the flawed meshes at geometrically complex regions to make room for Gaussians and refine the remaining ones at the texture-rich region to optimize the texture map. Then, we propose a transmittance-aware texture loss equipped with a warm-up to optimize the hybrid representation jointly. The overall pipeline of our approach is shown in Fig.~\ref{fig:pipeline}.

\subsection{Mesh Pruning}
Recent NeRF and Gaussian Splatting (GS) approaches extract meshes composed of numerous triangles from multi-view images. However, due to weak supervision in texture-less areas and challenges in representing complex geometries, the extracted meshes often exhibit significant geometric errors that adversely impact the hybrid representation. Furthermore, Gaussians outperform textured meshes in modeling geometrically complex and thin regions. To address these issues, we propose a pruning strategy that considers normal maps, adjacent angles, and mesh sizes to reduce the extracted mesh, thereby allocating space for Gaussian primitives to handle intricate regions.

We adopt the Planar-based Gaussian Splatting (PGSR)~\cite{DBLP:journals/corr/abs-2406-06521} as the baseline for initial mesh extraction. In texture-less regions, Gaussian primitives exhibit large scales $\boldsymbol{S}$, resulting in numerous mesh bumps with incorrect geometry that occlude surrounding objects. To mitigate this, we first prune the mesh by removing Gaussians exceeding a predefined scale threshold and subsequently extract the mesh using the truncated signed distance function (TSDF). This process yields a coarse mesh with minor bumps and rough surfaces, leading to increased disk storage and slower rendering speeds.

To further streamline the mesh, we apply a quadric error-based simplification method called QSlim to reduce the number of triangles to a desired count $K$, while preserving geometric topology. Next, we eliminate mesh segments representing complex geometries by analyzing the total variation of the normal map, angles between adjacent triangles, and triangle sizes. For each training view, we project the mesh onto 2D images to establish a mapping from pixels to triangles, enabling targeted pruning based on specific metrics.

Specifically, regions with high normal variation, indicating complex geometries, are identified using a prior normal map from the pre-trained StableNormal diffusion model. The top $\alpha_{normal}$ percent of pixels with the highest total variation are marked for triangle removal. Additionally, triangles with angles exceeding $45^{\circ}$ between adjacent faces are pruned to eliminate boundaries of flat regions or complex geometries. To remove trivial geometries that hinder texture map optimization as occluders, we sort and prune the smallest $\alpha_{area}$ percent of triangles based on their area sizes.

This comprehensive pruning process may result in floating triangles and tiny holes. To resolve these issues, we remove isolated connected triangles with counts below a minimum threshold $\alpha_{group}$ and close any resultant holes in the mesh to prevent floating geometries. Finally, we apply the LS3 subdivision surface algorithm to smooth the mesh further. All pruning and refinement operations are automated with predefined hyperparameters, resulting in a smooth mesh that predominantly represents texture-rich flat areas.

To incorporate high-frequency visual information from multi-view images, we attach texture coordinates (UV maps) to the mesh using Xatlas. We then initialize the texture map using Nviffrast, a differentiable rasterization tool, under image supervision. Consequently, we obtain a smooth and discrete mesh that effectively represents texture-rich flat regions, allowing us to gradually densify Gaussian primitives using the textured mesh as a background with fixed depth values.

\subsection{Joint optimization}
\paragraph{Joint rasterization}
As shown in Fig.~\ref{fig:pipeline}, we simultaneously optimized the 3DGS and the textured mesh in a differentiable way. For each training view, we first rendered the mesh to get a RGB image $\boldsymbol{I}_{m}$ and a depth map $\boldsymbol{D}_{m}$ and made the texture map optimizable. 
During the rasterization of GS, if the pixel is equipped with a triangle, we used the alpha-blending to produce an RGB value $\boldsymbol{I}_{gs}$ of Gaussian primitives before the triangle ($\boldsymbol{D}_{gs} \textless \boldsymbol{D}_{m}$) and got a transmittance coefficient $\boldsymbol{T}$ applied to the mesh. Otherwise, the rasterization performs as the vanilla GS. In short, we regard the mesh as an opaque background with a certain depth value. The final rasterization result is given as follows:
\begin{equation}
\nonumber
    \boldsymbol{I}_{h} = \left\{
\begin{aligned}
\boldsymbol{I}_{gs} & + \boldsymbol{T} \times \boldsymbol{I}_{m}, & \text{if triangle available} \\
\boldsymbol{I}_{gs} & + \boldsymbol{T} \times \boldsymbol{I}_{bg}, & \text{else} \\
\end{aligned}
\right.
\end{equation}
where $\boldsymbol{I}_{bg}$ denotes the predefined background color.
Following the training strategy of vanilla GS, we replaced the $\hat{\boldsymbol{I}}$ in eqn.~\ref{gs_loss} with the $\boldsymbol{I}_{h}$ to provide the supervision. 

\paragraph{Transmittance-aware supervision} Under the photometric supervision of the hybrid representation, the optimization process prefers to cover the front of the mesh with numerous Gaussian primitives. It will reduce the mesh's contribution to each pixel. 
To reduce the Gaussian primitives in front of the mesh, we adopt a $\lVert\boldsymbol{\widetilde{I}}, \boldsymbol{\widetilde{I}}_{m}\rVert_{2}^{2}$ norm loss between the masked ground truth  $\boldsymbol{\widetilde{I}}$ and the masked texture image $\boldsymbol{\widetilde{I}}_{m}$.  
$\boldsymbol{\widetilde{I}}$ equals to $\boldsymbol{I}$ when triangle is available for this pixel, else $\boldsymbol{\widetilde{I}}$ equals to $0$.

However, simply enforcing the rendered mesh to be close to the ground truth image will induce some ghosts on the mesh, shown in Fig.~\ref{fig: ghost}. When the explicit object in front of the back mesh is absent, in most cases in our settings, since we pruned the mesh with complex geometry. This supervision will project the object onto the back meshes along the ray from the camera center. When the mesh is viewed from other perspectives, we will see a ghost object that should not exist. First, we warm up the training only with the photometric loss for the first \textit{N} iterations to make up the absent object with Gaussians. Then we proposed a transmittance-aware texture loss $\mathcal{L}_{t}$ to prevent the incorrect supervision hindering the optimization of the texture map as follows:
\begin{equation}
    \mathcal{L}_{t}(\boldsymbol{I}, \boldsymbol{{I}}_{m}) = \mathcal{M}_{\boldsymbol{T}} \cdot \lVert\boldsymbol{\widetilde{I}}, \boldsymbol{\widetilde{I}}_{m} \rVert_{2}^{2},
\end{equation}
 $\mathcal{M}_{\boldsymbol{T}}$ denotes a transmittance-aware mask for each pixel. If the $\boldsymbol{T}$ is larger than 0.5, we prefer to rely on the textured mesh to dominate the pixel to reduce the Gaussian primitives. Otherwise, we hope to use the hybrid representation. Therefore, we used a \textit{sigmoid} function to get the mask for each pixel: 
 \begin{equation}
    \mathcal{M}_{\boldsymbol{T}}  = 1 / (1 + e^{-k \cdot (\boldsymbol{T} - 0.5)}).
\end{equation}

Since we adopt the $\mathcal{L}_{2}$ to reduce the Gaussian primitives and use hybrid representation to render the final image, we set $\lambda$  to zero after densification iteration (\emph{i.e} 15k) of 3DGS training. Combined with the warm-up strategy, some Gaussian primitives will be created at the object's location, making the transmittance for the back mesh less than $0.5$. Therefore, the object's color has a negligible impact on the optimization of back meshes. 
The joint optimization is supervised by: 
\begin{equation}
\label{eqn:total_loss}
\mathcal{L}(\boldsymbol{I}, \boldsymbol{I}_{h}, \boldsymbol{{I}}_{m}) = \mathcal{L}_{c}(\boldsymbol{I}, \boldsymbol{I}_{h}) + \lambda \cdot \mathcal{L}_{t}(\boldsymbol{I}, \boldsymbol{{I}}_{m}),
\end{equation}

where $\lambda$ controls the balance between the hybrid representation and the textured mesh.


\begin{table*}[htp]
    \centering

    \resizebox{2\columnwidth}{!}{
    \begin{tabular}{c|ccccc|ccccc}
        \toprule[1.1pt]
       \multirow{2}{*}{Method} &  \multicolumn{5}{c|}{Scannet++} & \multicolumn{5}{c}{Deep Blending} \\
       & SSIM $\uparrow$ & PSNR $\uparrow$ & LPIPS $\downarrow$ & \#Gaussian $\downarrow$  & FPS $\uparrow$ & SSIM $\uparrow$ & PSNR $\uparrow$ & LPIPS $\downarrow$ & \#Gaussian $\downarrow$  & FPS $\uparrow$  \\
       \midrule
       3DGS~\cite{DBLP:journals/tog/KerblKLD23}  & 0.862 & 24.22 & 0.248 & 0.911 M & 211 & 0.901 & 29.55 & 0.253 & 2.634 M &  346\\
       + Mesh (Ours) & 0.862 & 24.28 & 0.254 & \textbf{0.742M} & \textbf{231}  & 0.893 & 29.48 & 0.262 & \textbf{1.698 M} & \textbf{498}\\
       \midrule
       Mip-splatting~\cite{DBLP:conf/cvpr/YuCHS024}& 0.937 & 30.63 & 0.162 &	1.166 M & 164   &0.899	& 29.30& 0.248	&3.443M &129 \\
        + Mesh (Ours) & 0.934 & 30.46 & 0.168 &	\textbf{0.999 M} & \textbf{177}   & 0.893	& 29.23 & 0.259 & \textbf{2.251 M} & \textbf{176} \\
       \midrule
    Reduced GS~\cite{DBLP:journals/pacmcgit/PapantonakisKKLD24}  &  0.862& 24.09&0.248&0.427 M& 239  & 0.900 & 29.66 &	0.255 & 1.333 M & 223  \\ 
    + Mesh (Ours) & 0.861 & 24.09& 0.251 & \textbf{0.336 M} & \textbf{253} & 0.896 & 29.51& 0.260 & \textbf{0.668 M} & \textbf{317} \\
       \bottomrule[1.1pt]
    \end{tabular}
    }
    \caption {Quantitative evaluation comparing the proposed method with previous works on two indoor scenes. We report SSIM, PSNR , and LPIPS on test views, the number of Gaussians, and the FPS of the CUDA rasterizer routine.}
    \label{tab:quantitative}
\end{table*}

\section{Experiments}
We first present the details of the proposed hybrid representation. We then assess performance for indoor scenes on the Deep blending dataset~\cite{DBLP:journals/tog/HedmanPPFDB18} and the challenging Scannet++ dataset~\cite{DBLP:conf/iccv/YeshwanthLND23}.

\subsection{Implementation Details} 
We build our method upon the open-source 3DGS code. Following ~\cite{DBLP:journals/tog/KerblKLD23}, we train our models for 30K iterations across all scenes and use the same densification, schedule, and hyperparameters. We follow all the default settings of PGSR~\cite{DBLP:journals/corr/abs-2406-06521} to extract the initial mesh. We use the Nvdiffrast~\cite{Laine2020diffrast} to enable differentiable rendering of the mesh and the optimization of the textured map. Following ~\cite{DBLP:journals/pacmcgit/PapantonakisKKLD24}, the FPS measurements isolate the runtime of the CUDA rasterizer routine only and exclude any graphics API overheads. The final mesh contains about a hundred thousand triangles with a texture map of size $3 \times 2048 \times 2048$. 
We follow standard practice and report SSIM, PSNR, and LPIPS for rendering evaluation. All experiments are conducted on a single V100 GPU. More detailed information are provided in supplementary~\ref{sec: details}.

\paragraph{Datasets.} We verify the effectiveness of our approach using ten real-world indoor scenes from publicly available datasets: four scenes from the Deep blending~\cite{DBLP:journals/tog/HedmanPPFDB18} and six scenes from ScanNet++~\cite{DBLP:conf/iccv/YeshwanthLND23}. 


\subsection{Quantitative Results}
\paragraph{Rendering Performance.} The quantitative results of various methods are presented in Tab.~\ref{tab:quantitative}. We evaluated the standard metrics such as SSIM, PSNR, LPIPS, Gaussians number, and FPS, comparing against the baseline 3DGS and other Gaussians methods. Tab.~\ref{tab:quantitative} demonstrates that, in comparison to the 3DGS baseline, our hybrid representation effectively cuts down around 18\% and 35\% of Gaussian primitives used in 3DGS while maintaining comparable performance. 
We then incorporate the textured mesh into ~\cite{DBLP:journals/pacmcgit/PapantonakisKKLD24} with our joint optimization strategy to further decrease the number of the Gaussians by about 20\% and 50\% with a negligible impact on performance. Additionally, the real-time rendering speed is accelerated as a result of using fewer Gaussian primitives. We provide more detailed experimental results and analysis about integrating the mesh into GS methods in the supplementary~\ref{Sec: experiment}. 

\begin{figure*}
    \centering
    \includegraphics[width=1.0\linewidth]{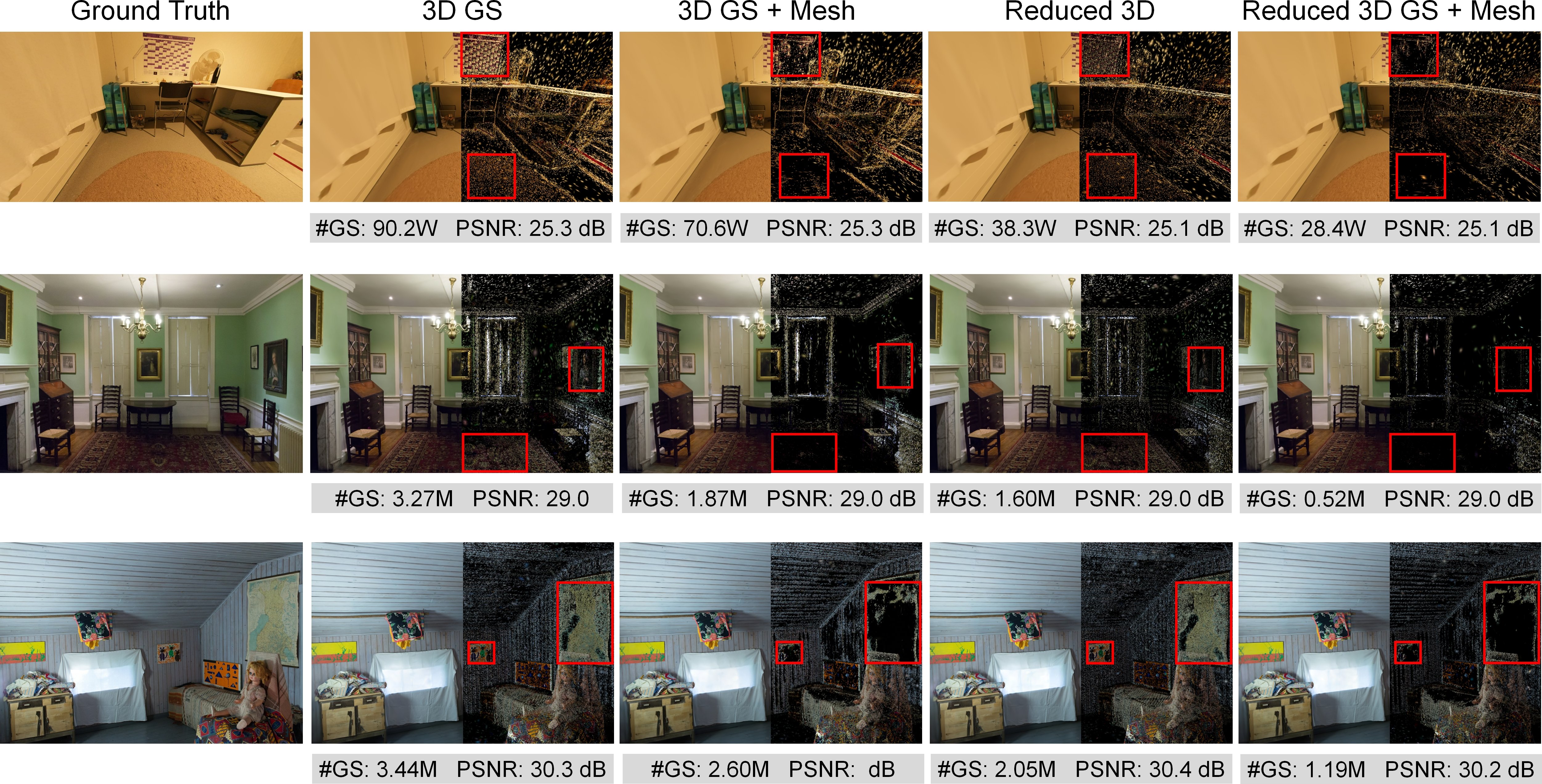}
    \caption{Visualization of the rendered images and Gaussian distribution of various methods. Our hybrid representation effectively reduces the Gaussians in texture-rich flat regions.}
    \label{fig:distribution}
\end{figure*}

Since Mip-splatting performs similarly to 3DGS, except for anti-aliasing, we demonstrate the rendered images and Gaussian distributions of 3DGS and Reduced GS in Fig.~\ref{fig:distribution}. the The hybrid representation can effectively reduce the Gaussians of the texture-rich regions.

\subsection{Qualitative Results}
\begin{figure*}
    \centering
    \includegraphics[width=1.0\linewidth]{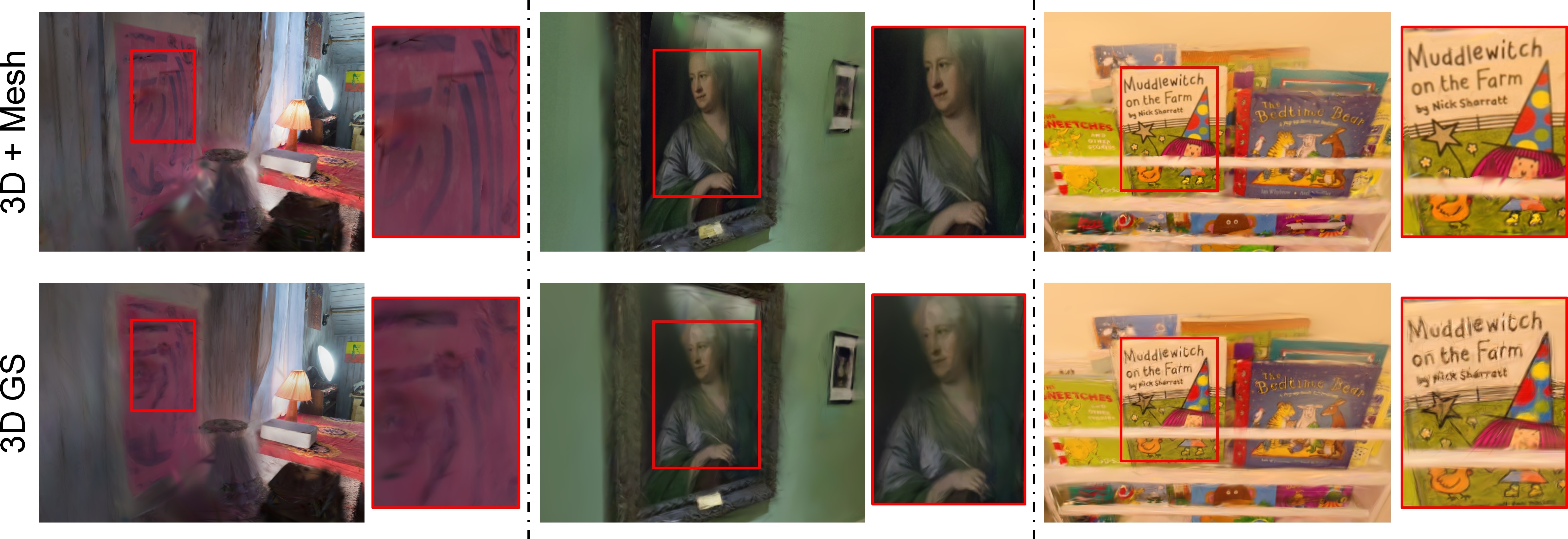}
    \caption{Comparison of the rendered images of 3DGS and our method. The hybrid representation demonstrates sharper renderings compared to the pure Gaussians, where the textured mesh delivers more precise background colors.}
    \label{fig: qualitative}
\end{figure*}

In addition to reducing the Gaussian primitives, the textured mesh demonstrates a sharper representation compared to the purely Gaussians. This is particularly evident in flat regions that are predominantly represented by the mesh.
The proposed transmittance-aware supervision plays a crucial role by directly translating the high-frequency visual details from the images into the texture map. As the visual comparisons illustrated in Fig.~\ref{fig: qualitative}, the mesh enhanced with a texture map is capable of rendering high-fidelity images, particularly in texture-rich flat regions. In contrast, rendering using Gaussian primitives tends to exhibit blurring and artifacts. This demonstrates the advantage of incorporating textured meshes in our hybrid approach to achieve superior visual quality in areas where detailed texture is essential. More visual results are provided in supplementary~\ref{sec:multiview}.

\subsection{Upper Bound}
To evaluate the upper bound of the hybrid representation and the proposed joint optimization strategy, we conduct experiments on the Replica dataset~\cite{DBLP:journals/corr/abs-1906-05797}, a synthesis indoor scene dataset with ideal meshes. 

We also apply the mesh prune onto the ideal mesh and then jointly optimize the GS and mesh to evaluate novel view renderings. Results in Tab.~\ref{tab:upper} show that if the mesh possesses an accurate geometry, the hybrid representation can achieve better performance, faster rendering with much fewer Gaussians.

\begin{figure}
    \centering
    \includegraphics[width=0.95\linewidth]{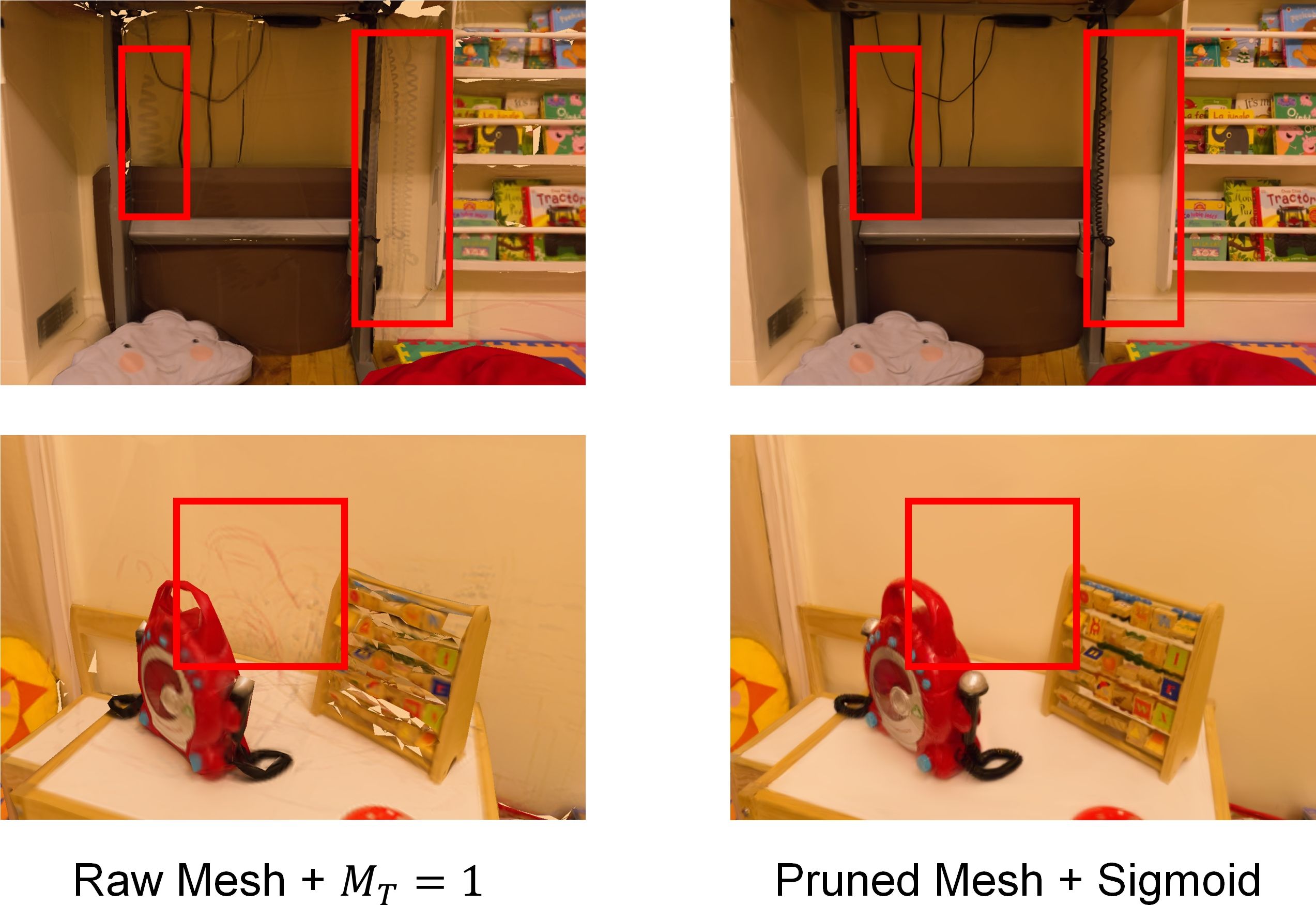}
    \caption{Illustration of incorrect color projection and the final renderings of our method.}
    \label{fig: ghost}
\end{figure}

\begin{table}[t]
    \centering
    \resizebox{1\columnwidth}{!}{
    \begin{tabular}{c|ccccc}
    \toprule[1.1pt]
    Method  & SSIM $\uparrow$ & PSNR $\uparrow$  & LPIPS $\downarrow$  & \#Gaussian $\downarrow$ & FPS $\uparrow$ \\
    \midrule
    3DGS   & 0.965 & 35.96 & 0.080 & 1.667 & 122 \\
    Ours   & 0.973 & 36.32 & 0.034 & 0.295 & 446 \\
    \bottomrule[1.1pt]
    \end{tabular}
    }
    \caption{Evaluation of hybrid representation on Replica.}
    \label{tab:upper}
\end{table}



\subsection{Ablation} 
We conduct detailed ablation studies using four indoor scenes from the deep bleending dataset to validate the key components of the proposed approach.

\paragraph{Mesh prune \& transmittance-aware supervision.}
We present the ablation study of our mesh prune strategy and transmittance-aware supervision in Tab.~\ref{tab:ablation}. For the raw mesh, we only use QSlim decimation to make it have the same number of faces as the mesh pruned by our method for a fare comparison. The results on deep bleeding demonstrate a marked improvement (around 1 dB) in rendering quality compared to the results of a naive combination of mesh and GS. For raw mesh without the transmittance-sensitive supervision, Gaussians were unable to compensate for errors caused by incorrect geometry in intricate parts (\textit{e.g.,} telephone toy) and missing local objects (\textit{e.g.,} wires) shown in Fig.~\ref{fig: ghost}. The object's color is wrongly projected onto the back mesh. The mesh pruning strategy can remove intricate parts, which leads to improved performance. We illustrate several visual comparisons of meshes before and after prune in supplementary~\ref{sec:mesh}.

We evaluate the effectiveness of the proposed transmittance-aware supervision on the pruned mesh. If $\mathcal{M}_{\boldsymbol{T}} $ is set to 1, there is an inability to distinguish between the foreground elements and the background components for each pixel of the training images. All pixel values are projected back to the mesh, resulting in inferior performance. The nonlinear sigmoid function performs better by acting as a soft indicator function.

\paragraph{Texture loss.} As illustrated in Tab.~\ref{tab:lambda}, we assess the balance between the photometric loss and the texture loss by varying the coefficient $\lambda$. Without the supervision between the mesh and ground truth images ($\lambda=0$), it will create more Gaussians in front of the mesh, thus weakening the representation of the textured mesh. Conversely, a high $\lambda$ will lead to a high transmittance to the mesh, which will make the transmittance-aware mask degrade to a constant $\mathcal{M}_{\boldsymbol{T}}=1$ mask. Besides, it forces images rendered solely from the textured mesh to overfit the ground truth and potentially impact the rendering of the hybrid representation. 



\begin{table}[ht]
    \centering
    
    \begin{tabular}{ccccc}
    \toprule[1.1pt]
    Mesh  & $\mathcal{M}_{\boldsymbol{T}}$ & SSIM $\uparrow$  & PSNR $\uparrow$ & LPIPS $\downarrow$ \\
    \midrule
    Raw   & 0 & 0.8812 & 28.52 & 0.2803\\
    Raw   & 1 & 0.8818 & 28.48 & 0.2806 \\
    \midrule
    Pruned   & 1 & 0.8913 & 29.31 &  0.2631\\
    Pruned   & T & 0.8915 & 29.41 & 0.2627\\
    Pruned   & Sigmoid & 0.8926 & 29.48 & 0.2627\\
    \bottomrule[1.1pt]
    \end{tabular}
    \caption{Comparison of raw and pruned mesh with different transmittance-aware mask.}
    \label{tab:ablation}
\end{table}

\begin{table}[ht]
    \centering

    \begin{tabular}{ccccc}
    \toprule[1.1pt]
    $\lambda$  & SSIM $\uparrow$ & PSNR $\uparrow$ & LPIPS $\downarrow$ & \#Gaussian $\downarrow$  \\
    \midrule
    0   & 0.891 & 29.42 & 0.263 & 1.875 M\\
    0.1 & 0.893 & 29.48 & 0.263 & 1.698 M\\
    0.2 & 0.891 & 29.40 & 0.263 & 1.678 M \\
    0.3 & 0.891 & 29.38 & 0.263 & 1.680 M\\
    \bottomrule[1.1pt]
    \end{tabular}
    \caption{Ablation study of the balance coefficient $\lambda$.} 
    \label{tab:lambda}
\end{table}

\begin{table}[ht]
    \centering

    \begin{tabular}{ccccc}
    \toprule[1.1pt]
    Method  & PSNR $\uparrow$   & \#Gaussian $\downarrow$  & Size $\downarrow$ \\
    \midrule
    0.5    & 29.24  & 1.715 M & 4.513 MB\\
    Xatlas  & 29.48  &  1.698 M & 6.012 MB\\
    Texrecon  & 29.50  & 1.715 M & 73.42 MB\\
    \bottomrule[1.1pt]
    \end{tabular}
    
    \caption{Comparison of different approaches to initialize the texture map. Size refers to the storage requirement of the texture map.} 
    \label{tab:tex_init}
\end{table}

\paragraph{Texture initialization.} We compare three methods for initializing the texture map. The first sets all values in the texture map to 0.5. The second uses Xatlas to get UV coordinates and then optimizes the texture map with Nvdiffrast. The third method utilizes texrecon~\cite{Waechter2014Texturing} to generate texture maps automatically. The first two methods allow setting the size of the texture map (set to $2024 \times 2024 \times 3$ as default), while texrecon generates multiple texture maps that need to be merged manually. The results are shown in Tab.~\ref{tab:tex_init}, we use the Xatlas as the default choice.
\section{Conclusion}
In this paper, we propose a hybrid representation that integrates textured meshes with 3D Gaussian Splatting (3DGS) to reduce the Gaussian primitives for indoor scenes. Based on the observation that texture-rich flat regions demand numerous Gaussian primitives to accurately capture significant color variations, we employ textured meshes to serve as background elements with specific depth values. This strategy substantially reduces the required number of Gaussians and accelerates rendering speeds. Additionally, we developed a mesh pruning method that considers normal maps, adjacent angles, and mesh sizes to eliminate meshes in geometrically intricate regions, thereby allocating space for Gaussian primitives. Our joint optimization technique ensures that both the textured mesh and Gaussian splats contribute effectively to the final representation, leveraging their respective strengths. Experimental results on the Deep Blending and ScanNet++ demonstrate that our hybrid approach maintains comparable rendering quality while significantly reducing the number of Gaussian primitives.

\paragraph{Limitations.} While our method excels in scenes with prominently texture-rich flat regions, it is less effective in environments dominated by multiple intricate structures unless a high-quality mesh is provided. Furthermore, our current approach does not optimize vertex coordinates or account for view-dependent effects of meshes. Addressing these limitations will be the focus of our future research, aiming to extend the applicability and robustness of our hybrid representation across a wider range of complex scenes.

\clearpage





\section*{Acknowledgments}
This work was supported in part by the Theme-based Research Scheme (TRS) project T45-701/22-R of the Research Grants Council (RGC), Hong Kong SAR.

\bibliographystyle{named}
\bibliography{ijcai25}

\clearpage
\appendix
\clearpage
\setcounter{page}{1}

\twocolumn[
\begin{@twocolumnfalse}
\section*{\centering{\textbf{\huge{Supplementary material}} \\[15pt]}}
\end{@twocolumnfalse}
]

\label{supp}
In this supplementary material, we provide additional implementation details experimental details in Sec.~\ref{sec: details}, the mesh comparisons before and after pruning in Sec~\ref{sec:mesh} and more  experimental results in Sec~\ref{Sec: experiment}. 

We offer a local web viewer incorporating videos of method comparison, which can be opened and viewed locally. (\textit{index.html} in the 
\textit{web\_viewer} folder.) Due to the 50MB space limit, we compressed the uploaded demo videos.

\section{Implementation details.} 
\label{sec: details}
\paragraph{Mesh Pruning.} To clear triangles in weak textured areas with large geometric errors, we remove the big Gaussian primitives before extracting mesh from PGSR. We calculated the camera extent $D$ defined in 3DGS and deleted all the large Gaussians whose maximum axis length is larger than $0.01\times D$. Then, we extracted the raw mesh using the truncated signed distance function (TSDF) with the voxel size set to 0.01 and maximum depth set to 100. For mesh simplification, the number $K$ after QSlim,  $\alpha_{normal}$, $\alpha_{area}$, and $\alpha_{group}$ are set to $500K, 20, 50, 100$ as default. The balance coefficient $\lambda$ and warm-up \textit{N} iterations for joint optimization are set to $0.1$ and $3k$. The $k$ in \textit{sigmoid} function of transmittance-aware mask is set to $20$ as default.

\paragraph{Algorithm.}
The overall algorithm is shown in Alg.~\ref{alg1:overall}.

\begin{algorithm}[!ht]
\vspace{5pt}
\KwInput{Images $\boldsymbol{I}$, Mesh $\mathcal{M}$}
\KwOutput{3DGS $\mathcal{GS}$, Mesh $\mathcal{M}$, Texture map $\mathcal{T}$}

\vspace{5pt}
\textbf{Stage 1: Mesh pruning} \\
\hspace{0.3em} $\text{Normal}: \boldsymbol{M}_{N} = \textit{StableNormal} (\boldsymbol{I}, \alpha_{normal}) $ \;
\hspace{0.3em} $\text{Angle}: \boldsymbol{M}_{A} = \textit{AdjacentAngle} (\mathcal{M}, 45\text{\textdegree}) $ \;
\hspace{0.3em} $\text{Area size}: \boldsymbol{M}_{S} = \textit{AreaSize} (\mathcal{M}, \alpha_{area}) $ \;
\hspace{0.3em} $\text{Prune mask}: \boldsymbol{M}_{P} = \boldsymbol{M}_{N} \text{\,\&\&\,} \boldsymbol{M}_{A} \text{\,\&\&\,} \boldsymbol{M}_{S}$ \;
\hspace{0.3em} $\text{Pruned Mesh}: \mathcal{M_{P}}=\textit{Smooth}(\textit{Prune}(\mathcal{M},\boldsymbol{M}_{P}))$ \;
\hspace{0.3em} $\text{Texture map}: \mathcal{T}=\textit{Nvdiffrast}(\textit{Xatlas}(\mathcal{M_{P}}), \boldsymbol{I})$ \;
\vspace{5pt}
\textbf{Stage 2:Joint optimization} \\
\hspace{0.3em} $ \boldsymbol{{I}}_{m}, \boldsymbol{{D}}_{m} = \textit{Nvdiffrast}(\mathcal{M_{P}}, \mathcal{T})$ \;
\hspace{0.3em} $ \boldsymbol{{I}}_{h} = \textit{Rasterize}(\mathcal{GS},\boldsymbol{{I}}_{m}, \boldsymbol{{D}}_{m}) $ \;
\hspace{0.3em} \For{$i=0, 1, ..., \text{MaxIter}$}{
    \eIf{$i > i_{warm-up}\,\&\&\, i < i_{densification}$}
    {
    $\mathcal{L}(\boldsymbol{I}, \boldsymbol{I}_{h}, \boldsymbol{{I}}_{m})\rightarrow \text{Update}(\mathcal{T}, \mathcal{GS})$\;
    }
    {
    $\mathcal{L}_c(\boldsymbol{I}, \boldsymbol{I}_{h})\rightarrow \text{Update}(\mathcal{T}, \mathcal{GS})$\;
    }
    
}
\caption{The overall algorithm.}
\label{alg1:overall}
\end{algorithm}

\paragraph{Datasets.} We used four indoor scenes \textit{playroom, drjohnson, bedroom, creepyattic} of deep blending dataset~\cite{DBLP:journals/tog/HedmanPPFDB18} and six indoor scenes \textit{07ff1c45bb, 0a184cf634, 0a7cc12c0e, 0cf2e9402d, 1366d5ae89, 1a8e0d78c0} of Scannet++~\cite{DBLP:conf/iccv/YeshwanthLND23} to evaluate the effectiveness of our approach.  For deep bleeding, we select one out of every eight pictures to form the test dataset. Regarding Scannet++, we adhere to the predefined file to split the dataset. All metrics are evaluated on the test dataset.

\section{Mesh comparison}
\label{sec:mesh}
\begin{figure*}
    \centering
    \includegraphics[width=1.0\linewidth]{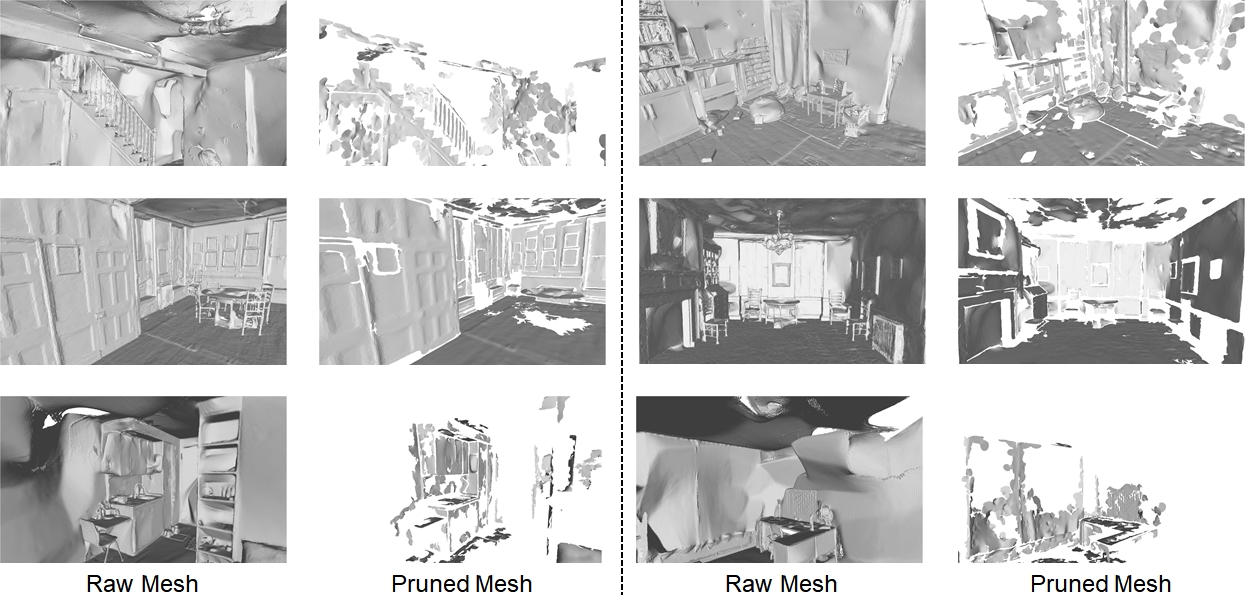}
    \caption{Illustration of mesh before and after pruning.}
    \label{fig: mesh comparison}
\end{figure*}

As illustrated in Fig.~\ref{fig: mesh comparison}, the extracted mesh exhibits significant geometric errors, particularly in regions with weak texture (\textit{e.g.,} ceiling, walls) and strong geometry (\textit{e.g.,} stroller, radiator). The pruned mesh, however, is capable of substantially clearing the mesh in such areas and performing a certain level of smoothing. Nevertheless, even after pruning, there remain regions within the mesh that possess geometric errors. Enhancing the accuracy of the mesh can further contribute to the overall performance improvement.

\section{More Experiments.} 
\label{Sec: experiment}
\subsection{Analysis.}

For each pixel, when the mesh is available, the difference between the mesh-rendered image and the ground truth is less than that of the randomly initialized Gaussians-rendered image and the ground truth. This results in a small gradient of the Gaussian's position, restraining the clone and split operation and reducing the number of Gaussian primitives. Intuitively, a correct color from the mesh-rendered image implies no necessity to densify more Gaussians for compensating the photometric difference.


\paragraph{Per-scene Metrics.}
We provide SSIM, PSNR, LPIPS, the number of Gaussian primitives and FPS for ten scenes in Tab.~\ref{tab:SSIM}-\ref{tab:FPS}, including four from deep bleeding and six from Scannet++.

\begin{table*}[htp]
    \centering
    \caption {SSIM scores for ten scenes.}
    \resizebox{2\columnwidth}{!}{
    \begin{tabular}{c|cccccc|cccc}
        \toprule[1.1pt]
        \multirow{2}{*}{Method} &  \multicolumn{6}{c|}{Scannet++} & \multicolumn{4}{c}{Deep Blending} \\
          & 07ff1c45bb  & 0a184cf634 & 0a7cc12c0e  & 0cf2e9402d & 1366d5ae89 & 1a8e0d78c0 & Playroom & Drjohnson & Bedroom  & Creepyattic  \\
    \midrule

    3DGS~\cite{DBLP:journals/tog/KerblKLD23} & 0.842 & 0.885 &0.879 &0.839 &0.856 & 0.870  & 0.908 & 0.900 & 0.905 & 0.889\\
     + Mesh (Ours) & 0.842 &0.887 &0.879 &0.833 &0.857 &0.873 & 0.901 & 0.888 & 0.902 & 0.880  \\
     \midrule
    Mip-splatting~\cite{DBLP:conf/cvpr/YuCHS024}& 0.922 &0.962 &0.959 &0.928 &0.928 &0.922 & 0.906 & 0.897 & 0.906 & 0.887\\
    + Mesh (Ours) & 0.917 & 0.962 &0.956 &0.924 &0.924 & 0.920 &0.906 & 0.887 & 0.900 & 0.880\\
    \midrule
    Reduced GS~\cite{DBLP:journals/pacmcgit/PapantonakisKKLD24} & 0.842 &0.887 &0.878 &0.837 &0.856 &0.872 & 0.903 & 0.901 & 0.908 & 0.887\\ 
    + Mesh (Ours) & 0.840 &0.886 &0.876 &0.837 &0.854 &0.871 & 0.901 & 0.900 & 0.901 & 0.880\\
    \bottomrule[1.1pt]
    \end{tabular}
    \label{tab:SSIM}
    }
\end{table*}

\begin{table*}[htp]
    \centering
    \caption {PSNR scores for ten scenes.}
    \resizebox{2\columnwidth}{!}{
    \begin{tabular}{c|cccccc|cccc}
        \toprule[1.1pt]
        \multirow{2}{*}{Method} &  \multicolumn{6}{c|}{Scannet++} & \multicolumn{4}{c}{Deep Blending} \\
      & 07ff1c45bb  & 0a184cf634 & 0a7cc12c0e  & 0cf2e9402d & 1366d5ae89 & 1a8e0d78c0 & Playroom & Drjohnson & Bedroom  & Creepyattic  \\
    \midrule

    3DGS~\cite{DBLP:journals/tog/KerblKLD23} & 23.66&24.79& 25.31&22.53&24.39&24.67 & 29.98 & 28.96 & 28.93 & 30.33
\\
     + Mesh (Ours) & 23.64&24.94&25.31&22.39&24.66&24.83 & 30.11 & 29.01 & 28.69 & 30.11
\\
\midrule
    Mip-splatting~\cite{DBLP:conf/cvpr/YuCHS024}& 28.73&33.96&33.77&29.59&29.19&28.56 & 30.11 & 28.22 & 28.75 & 30.15
\\
    + Mesh (Ours) & 28.51 & 33.73 & 33.65 & 29.39 & 28.96 & 28.51 & 30.08 & 28.46 & 28.50 & 29.92 \\
\midrule
    Reduced GS~\cite{DBLP:journals/pacmcgit/PapantonakisKKLD24} & 23.62&24.74 &25.10 &22.00 &24.48& 24.61 & 30.12 & 29.07 & 28.99 & 30.44
  \\ 
     + Mesh (Ours) & 23.58&24.82&25.05&22.24&24.42&24.44 & 30.16 & 29.01 & 28.73 & 30.18
 \\
    \bottomrule[1.1pt]
    \end{tabular}
    \label{tab:PSNR}
    }
\end{table*}

\begin{table*}[htp]
    \centering
    \caption {LPIPS scores for ten scenes.}
    \resizebox{2\columnwidth}{!}{
    \begin{tabular}{c|cccccc|cccc}
        \toprule[1.1pt]
        \multirow{2}{*}{Method} &  \multicolumn{6}{c|}{Scannet++} & \multicolumn{4}{c}{Deep Blending} \\
      & 07ff1c45bb  & 0a184cf634 & 0a7cc12c0e  & 0cf2e9402d & 1366d5ae89 & 1a8e0d78c0 & Playroom & Drjohnson & Bedroom  & Creepyattic  \\
    \midrule

    3DGS~\cite{DBLP:journals/tog/KerblKLD23} & 0.256 &0.246&0.244&0.269&0.236&0.239  & 0.249 & 0.250 & 0.282 & 0.231
\\
    + Mesh (Ours) & 0.268&0.250&0.247&0.282&0.238&0.241 & 0.255 & 0.262 & 0.290 & 0.243 \\
    \midrule
    Mip-splatting~\cite{DBLP:conf/cvpr/YuCHS024}& 0.162&0.164&0.155&0.173 & 0.157&0.162&0.240 & 253 & 0.277 & 0.224
\\
    + Mesh (Ours) & 0.171 & 0.167 & 0.158 & 0.182 & 0.164 & 0.166  & 0.247 & 0.266 & 0.289 & 0.234 \\
    \midrule
    Reduced GS~\cite{DBLP:journals/pacmcgit/PapantonakisKKLD24} & 0.256& 0.245&0.244&0.269&0.237&0.238 & 0.253 & 0.250 & 0.282 & 0.236
 \\ 
     +  Mesh (Ours) &  0.262&0.249&0.246&0.272&0.234&0.240 & 0.252 & 0.255 & 0.286 & 0.254 \\
    \bottomrule[1.1pt]
    \end{tabular}
    \label{tab:LPIPS}
    }
\end{table*}

\begin{table*}[htp]
    \centering
    \caption {The number of Gaussian primitives for ten scenes.}
    \resizebox{2\columnwidth}{!}{
    \begin{tabular}{c|cccccc|cccc}
        \toprule[1.1pt]
        \multirow{2}{*}{Method} &  \multicolumn{6}{c|}{Scannet++} & \multicolumn{4}{c}{Deep Blending} \\
      & 07ff1c45bb  & 0a184cf634 & 0a7cc12c0e  & 0cf2e9402d & 1366d5ae89 & 1a8e0d78c0 & Playroom & Drjohnson & Bedroom  & Creepyattic  \\
    \midrule

    3DGS~\cite{DBLP:journals/tog/KerblKLD23} & 1.434 M & 0.577 M & 0.902 M & 0.913 M & 0.776 M & 0.867 M & 2.145 M& 3.275 M& 1.682 M& 3.436 M\\
    + Mesh (Ours) & 1.086 M & 0.452 M & 0.706 M & 0.762 M & 0.670 M & 0.777 M & 1.502 M& 1.853 M& 0.941 M& 2.495 M\\
    \midrule
    Mip-splatting~\cite{DBLP:conf/cvpr/YuCHS024}& 1.785 M & 0.750 M & 1.105 M & 1.269 M & 1.007 M & 1.082 M & 2.869 M& 4.183 M& 1.958 M& 4.764 M\\
    + Mesh (Ours) &  1.475 M & 0.640 M & 0.899 M & 1.101 M & 0.847  M & 1.033 M &2.067 M &2.155 M &1.221 M &3.564 M \\
    \midrule
    Reduced GS~\cite{DBLP:journals/pacmcgit/PapantonakisKKLD24} &  0.645 M & 0.281 M & 0.383 M & 0.484 M & 0.371 M & 0.398 M & 0.965 M& 1.602 M& 0.709 M& 2.055 M\\ 
    + Mesh (Ours) & 0.466 M & 0.222 M & 0.256 M & 0.434 M & 0.305 M & 0.334 M & 0.590 M & 0.495 M & 0.341 M & 1.194 M \\
    \bottomrule[1.1pt]
    \end{tabular}
    \label{tab:GS}
    }
\end{table*}

\begin{table*}[htp]
    \centering
    \caption {FPS for ten scenes.}
    \resizebox{2\columnwidth}{!}{
    \begin{tabular}{c|cccccc|cccc}
        \toprule[1.1pt]
        \multirow{2}{*}{Method} &  \multicolumn{6}{c|}{Scannet++} & \multicolumn{4}{c}{Deep Blending} \\
      & 07ff1c45bb  & 0a184cf634 & 0a7cc12c0e  & 0cf2e9402d & 1366d5ae89 & 1a8e0d78c0 & Playroom & Drjohnson & Bedroom  & Creepyattic  \\
    \midrule

    3DGS~\cite{DBLP:journals/tog/KerblKLD23} & 173 & 277 & 212 & 162 & 207 & 236 & 181 & 392 & 662 & 148\\
    + Mesh (Ours) & 204 & 298 & 234 & 198 & 214 & 239 & 239 & 649 & 854 & 250\\
    \midrule
    Mip-splatting~\cite{DBLP:conf/cvpr/YuCHS024}& 136 & 216 & 169 & 113 & 174 & 175 & 136 & 99 & 163 & 121\\
    + Mesh (Ours) & 150 & 229 & 176 & 126 & 188 & 196 & 167 & 186 & 134 & 218 \\
    \midrule
    Reduced GS~\cite{DBLP:journals/pacmcgit/PapantonakisKKLD24} &  219 & 290 & 245 & 178 & 249 & 254 & 235 & 194 & 270 & 193 \\ 
     + Mesh (Ours) & 236 & 302 & 269 & 192 & 255 & 263 & 319 & 324 & 343 & 282\\
    \bottomrule[1.1pt]
    \end{tabular}
    \label{tab:FPS}
    }
\end{table*}

\subsection{More Qualitative Results}
\label{sec:multiview}
\begin{figure*}
    \centering
    \includegraphics[width=1.0\linewidth]{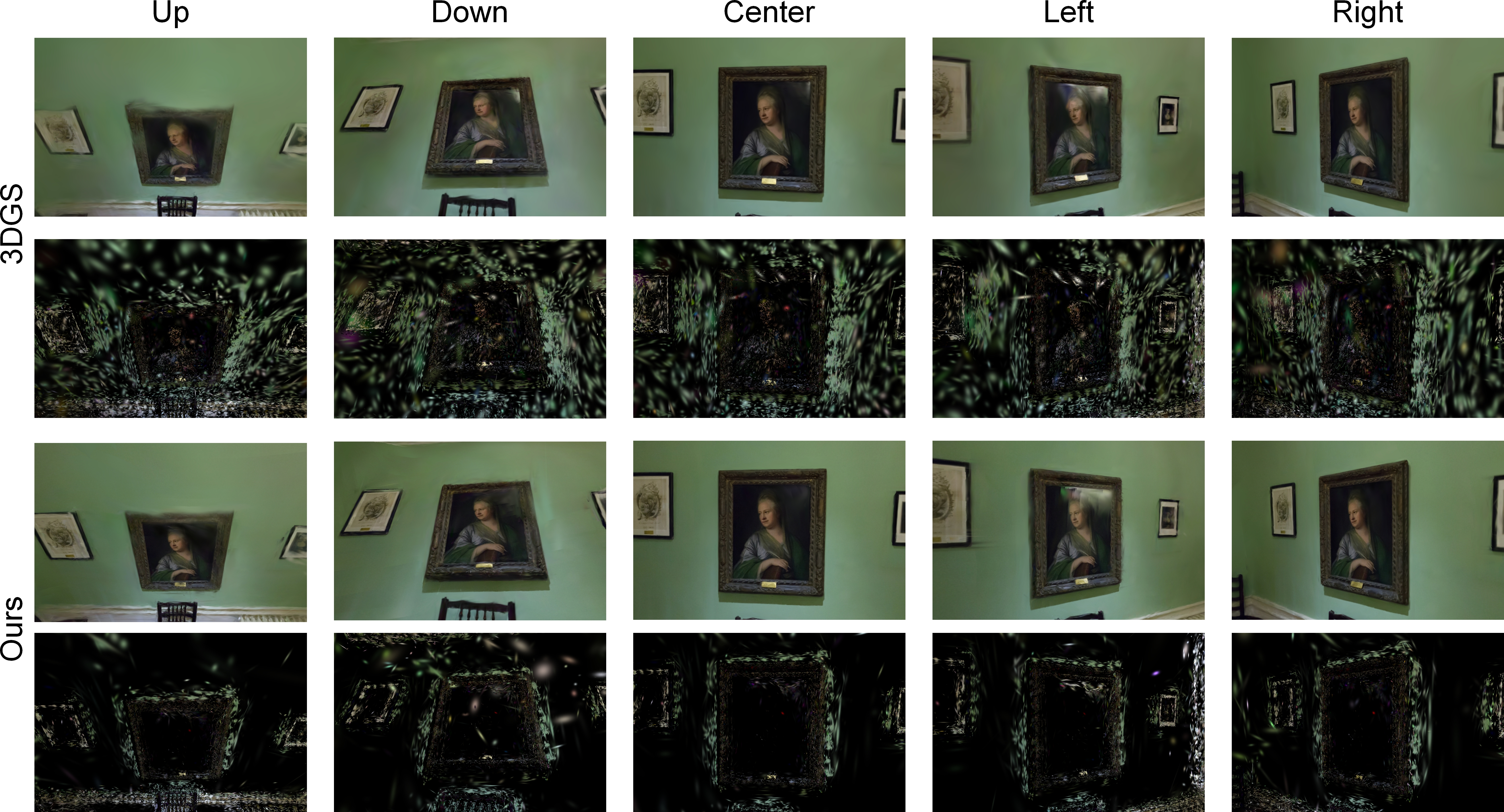}
    \caption{Rendered images and Gaussian distribution from five perspectives of Drjohnson.}
    \label{fig: view_drjohnson}
\end{figure*}

\begin{figure*}
    \centering
    \includegraphics[width=1.0\linewidth]{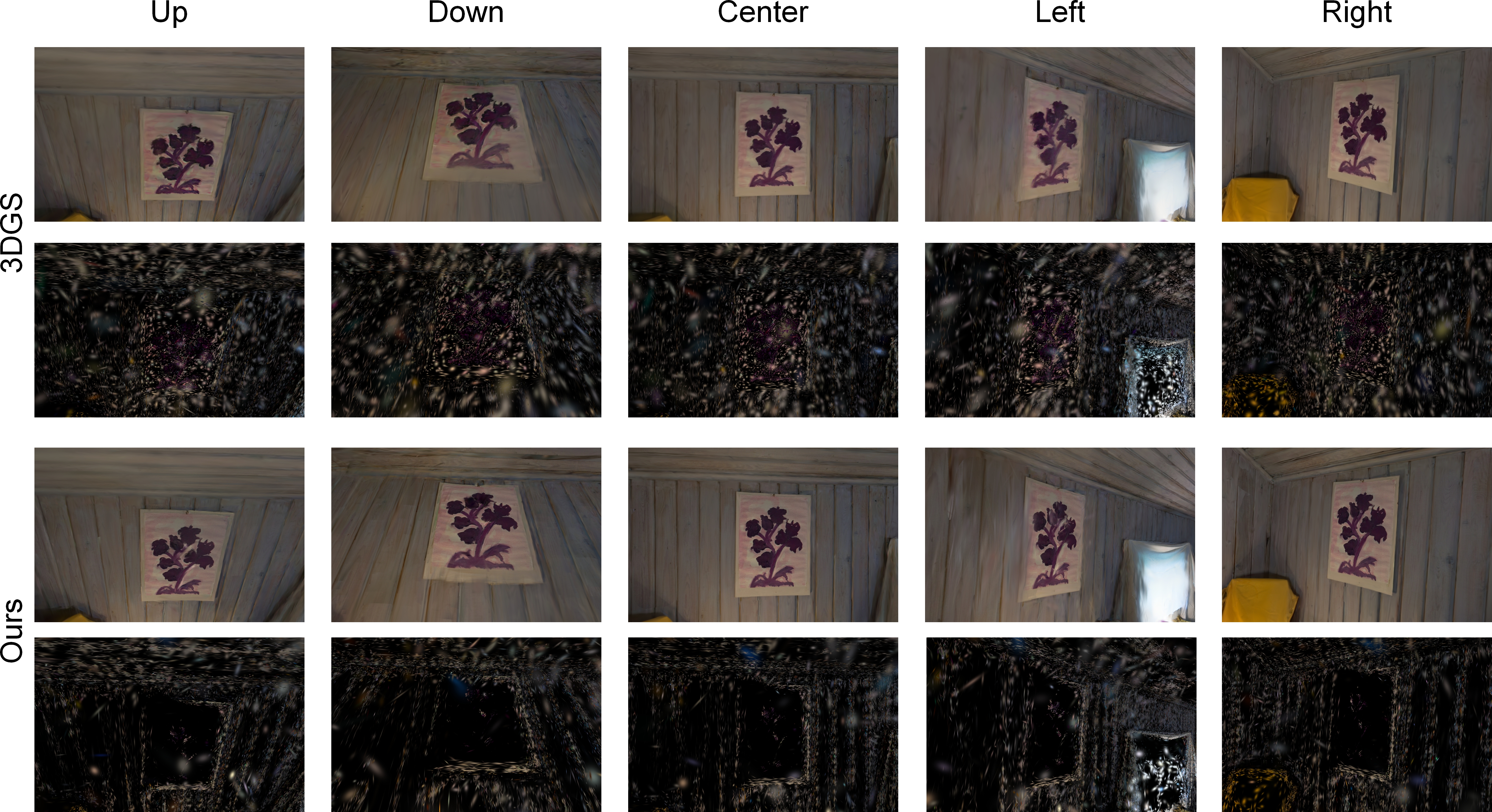}
    \caption{Rendered images and Gaussian distribution from five perspectives of Creepyattic.}
    \label{fig: view_creepyattic}
\end{figure*}

\paragraph{Multiple Perspectives.} We rendered the images and Gaussian distributions of regions where the mesh is available from five perspectives, namely up, down, left, right, and center.  As illustrated in Fig.~\ref{fig: view_drjohnson} and Fig.~\ref{fig: view_creepyattic}, the rendered images of our hybrid representation exhibit consistency and sharpness across multiple viewpoints with less Gaussian primitives. In contrast, 3DGS generates some blurs when the views are outside the training trajectory (such as the top, down, and left perspectives). 

\paragraph{Video Comparison.}
To visually compare the rendering results of 3DGS and our method, we recorded rendering results along the same camera trajectory for visualization. Meanwhile, to compare the number of Gaussian primitives, we scaled down all Gaussians by a factor of five and visualized their distribution. Results are stored in \textit{index.html} of the supplementary material.

\end{document}